\documentclass[conference]{IEEEtran}
\IEEEoverridecommandlockouts

\usepackage{multirow,graphicx}
\usepackage{threeparttable}
\usepackage{float}
\usepackage{paralist}
\usepackage{balance}
\usepackage{hyperref}
\usepackage{cite}
\usepackage{amsmath,amssymb,amsfonts}
\usepackage{algorithmic}
\usepackage{graphicx}
\usepackage{textcomp}
\usepackage{xcolor}
\graphicspath{ {img/} }
\def\BibTeX{{\rm B\kern-.05em{\sc i\kern-.025em b}\kern-.08em
    T\kern-.1667em\lower.7ex\hbox{E}\kern-.125emX}}
\begin{document}

\title{People Tracking and Re-Identifying in Distributed
Contexts: Extension Study of PoseTReID}

\makeatletter
\newcommand{\linebreakand}{%
  \end{@IEEEauthorhalign}
  \hfill\mbox{}\par
  \mbox{}\hfill\begin{@IEEEauthorhalign}
}
\newcommand\myeq{\mkern1.5mu{=}\mkern1.5mu}
\makeatother
\author{
\IEEEauthorblockN{Ratha Siv}
\IEEEauthorblockA{University of Mons}
\and
\IEEEauthorblockN{Matei Mancas}
\IEEEauthorblockA{University of Mons}
\and
\IEEEauthorblockN{Bernard Gosselin}
\IEEEauthorblockA{University of Mons}
\linebreakand 
\IEEEauthorblockN{Dona Valy}
\IEEEauthorblockA{Institute of Technology of Cambodia}
\and
\IEEEauthorblockN{Sokchenda Sreng}
\IEEEauthorblockA{Institute of Technology of Cambodia}
}

\maketitle

\begin{abstract}
In our previous paper, we introduced PoseTReID which is a generic framework for real-time 2D multi-person tracking in distributed interaction spaces where long-term people's identities are important for other studies such as behavior analysis, etc. In this paper, we introduce a further study of PoseTReID framework in order to give a more complete comprehension of the framework. We use a well-known bounding box detector YOLO (v4) for the detection to compare to OpenPose which was used in our last paper, and we use SORT and DeepSORT to compare to centroid which was also used previously, and most importantly for the re-identification, we use a bunch of deep leaning methods such as MLFN, OSNet, and OSNet-AIN with our custom classification layer to compare to FaceNet which was also used earlier in our last paper. By evaluating on our PoseTReID datasets, even though those deep learning re-identification methods are designed for only short-term re-identification across multiple cameras or videos, it is worth showing that they give impressive results which boost the overall tracking performance of PoseTReID framework regardless the type of tracking method. At the same time, we also introduce our research-friendly and open source Python toolbox pyppbox, which is purely written in Python and contains all sub-modules which are used in this study along with real-time online and offline evaluations for our PoseTReID datasets. This pyppbox is available on GitHub \url{https://github.com/rathaumons/pyppbox} .
\end{abstract}


\section{Introduction}

Previously in our last paper \cite{ptreid9271712}, we introduced PoseTReID framework which was designed for both short-term and long-term people tracking in distributed people interaction spaces or various ecological applications either indoor like in malls, outdoor like in smart cities, or both like in amusement parks. We used OpenPose \cite{cao2017realtime} for people detection, centroid tracker for people tracking, and FaceNet \cite{schroff2015facenet} for both short-term and long-term people re-identification. We also introduced PoseTReID datasets \cite{ptreid9271712} which were used for evaluating the results. The overall tracking performance results of PoseTReID in our last paper outperformed the state-of-art tracker-only STAF \cite{raaj2019efficient_staf} in almost every scenario except in the surveillance video where the camera's position is a bit too high and too far for FaceNet to re-identify. 

\begin{figure}[ht]
  \centering
  \includegraphics[width=\linewidth]{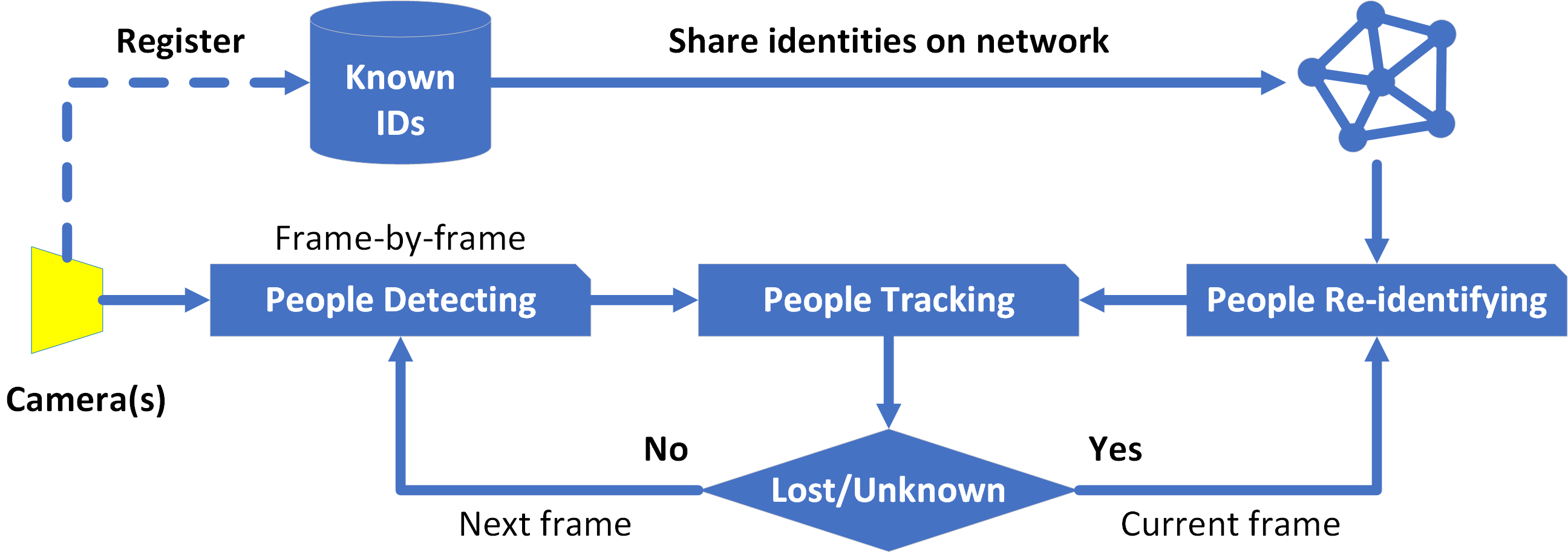}
  \caption{The generic architecture of PoseTReID framework: people detecting, people tracking, and people re-identifying modules work together frame by frame where the re-identifying module helps re-identify When the tracking is lost or finds an unknown.}
  \label{fig:res}
\end{figure}

As an extension of our last paper, in this paper, we give a more complete comprehensive study of PoseTReID framework and show the potential of the overall tracking performance with a selection of different detector, tracker, and re-identifier modules with some other improvements we made. The details are described in the next section.

\section{Experimental Study}

We still stick with our context like we mentioned in our previous paper; there is basically a need for (1) long-term people re-identification and (2) body features and poses for interactions. However, in this paper, we give a more complete and general comprehensive study as an extension of our previous work by comparing various detectors, trackers, and re-identifiers which are described in this section.

\subsection{Detector}

\subsubsection{OpenPose} 

OpenPose \cite{cao2017realtime} is a real-time deep learning-based multi-person system that can jointly detect human body, hands, feet as keypoints, as well as facial landmarks. With a dedicated GPU, OpenPose can perform multi-person detection on RGB streams in real-time. OpenPose only produces frame by frame detection, and it does not have any official or stable built-in tracker yet. 

\subsubsection{YOLO}

YOLO (You Only Look Once) \cite{yolo} is a fast real-time object detector which uses only a single neural network to detect the whole image. The network divides input image into small regions and then predicts classes and gives them bounding boxes and probabilities. The original authors kept improving until v3 \cite{yolov3}, and we started to see different authors from version v4 \cite{yolov4}.

Our Python toolbox pyppbox uses OpenCV dnn (Deep Neural Network) module \cite{opencv_library} which supports most versions of YOLO. However, in this paper, we only include v4 \cite{yolov4} of YOLO for our comparison study.

\subsection{Tracker}

\subsubsection{Centroid}

Centroid or Euclidean distance is a simple, yet highly effective object tracking. It was implemented in our last paper on top of OpenPose where neck was chosen as the represent point of every detected person because neck point is one of the most stable and visible nodes given by OpenPose detection.

\begin{figure}[ht]
  \centering
  \includegraphics[width=\linewidth]{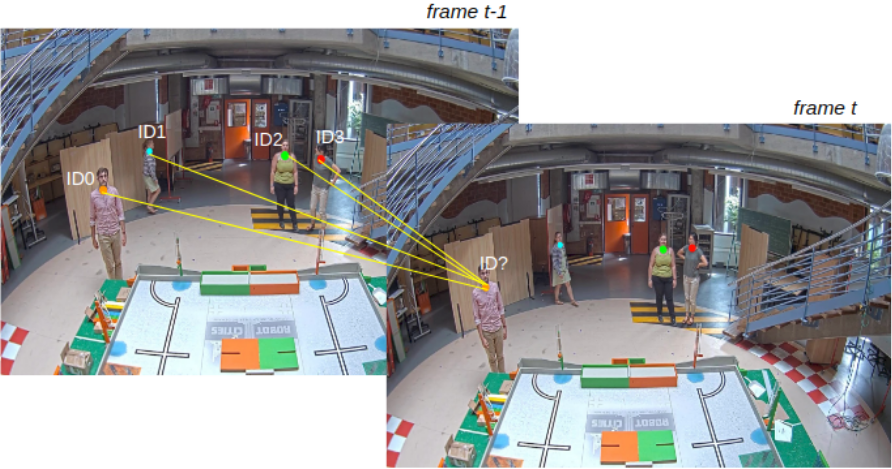}
  \caption{Illustration of the centroid tracking algorithm. We compute the Euclidean distances between each pair of original centroids at frame {\textit{t-1}} and new centroids at frame {\textit{t}}. IDs are matched by the closest distances.}
  \label{fig:tracking}
\end{figure}

\begin{equation}
\forall P_{i,t}, ID_{i,t} \gets ID_{j,t-1} \mid ID_{j,t-1}\myeq f_{t}(min(D))\label{eq}
\end{equation}

For every detected person at a frame {\textit{t}}, we compute the Euclidean distance between each pair of existing centroids (necks), $D\myeq \{\,\lVert d_{j}\lVert\,\,\,\mid\lVert d_{j}\lVert\myeq d(P_{i,t}, P_{j,t-1})\}$.

\subsubsection{SORT}

SORT (Simple Online and Realtime Tracking) \cite{sort} is a well-known open source of a visual multiple object tracking framework based on rudimentary data association and state estimation techniques. It is a real-time and online bounding box tracker where requires only previous and current inputs from the bounding detector. SORT performs well, but it does not handle occlusion or re-identify object after it disappears and reappears.

We add SORT to our comparison study to see how it performs with other modules in our PoseTReID framework.

\subsubsection{DeepSORT}

DeepSORT (Simple Online and Realtime Tracking with a Deep Association Metric) \cite{deepsort} is designed as an upgrade over SORT, which integrates a deep association metric as an internal re-identifier in the aim of solving occlusion and reducing the number of identity switches. However, DeepSORT is much computational heavier than SORT.

We also add DeepSORT to our comparison study to see how it performs with other modules in our PoseTReID framework.

\subsection{Re-Identifier}

In our PoseTReID framework, the re-identifier has 3 rules \cite{ptreid9271712} for handling occlusion and duplicated ID as follows:
\begin{itemize}
  \item When a new or unknown person is detected, meaning the tracking needs an ID, re-identifying (ReID) is applied to this person only.
  \item When the same ID is allocated or found on multiple persons, re-identifying (ReID) is applied again on all people having the same ID.
  \item A speed limit threshold might be or be not used: When the distance between frames of the same ID is greater than the limit, ReID is applied again. This last rule is not used in this paper.
\end{itemize}

\subsubsection{FaceNet}
\label{sssec:facenet}

FaceNet \cite{schroff2015facenet} is an end-to-end learning which learns to create euclidean space by mapping faces, and the face similarity can be measured by the corresponded distances. The model needs to be first trained on large datasets such as CASIA-WEBFACE \cite{yi2014learning}, MS-Celeb-1M \cite{guo2016ms}, VGGFace2 \cite{cao2018vggface2}, etc. 

We use pre-trained VGGFace2 model for training the characters in our PoseTReID datasets. In our previous paper, our PoseTReID with the combination of OpenPose as detector, Centroid as tracker, and FaceNet as re-identifier, outperformed others in almost all scenarios except the Hard-Surveillance complexity where FaceNet was struggle to re-identify the characters. We looked into that issue, and we later identified that our classification model did not perform well enough. This time, we tweak our classification model by changing the SVM \cite{708428_svm} kernel $k(x,x')$ function from linear to radial basis function:

\begin{equation}
k(x,x') = \exp(\frac{-(x-x')^2}{2\gamma^2})
\end{equation}

Where $\gamma$ is the length scale of the kernel.

We also add more face data of every character for training our classifier model (About 600 faces per character and approximately 2,957 images in total for the 5 characters in our PoseTReID datasets).

\subsubsection{MLFN}

MLFN \cite{mlfn} proved that the effectiveness of person re-identification can be achieved by modeling discriminative and view-invariant factors of person appearance at both high and low semantic levels. The authors proposed Multi-Level Factorisation Net which models the visual appearance of a person into latent discriminative factors at multiple semantic levels without extra manual annotation. MLFN achieves state-of-the-art results on many popular datasets.

We use the pre-trained MLFN model (Provided by torchreid \cite{torchreid}) with our custom classification layer for training the characters in our PoseTReID datasets. Similar to FaceNet module, we extract 30 images of full body crop of all characters, and then train with the same SVM configuration.

\subsubsection{OSNet}

OSNet \cite{osnet} is a lightweight yet effective deep learning person re-identification model. Similar to MLFN, OSNet also relies on discriminative features which capture different spatial scales and encapsulate an arbitrary combination of multiple scales which are addressed as homogeneous and heterogeneous scales omni-scale features by the authors. Omni-Scale Network (OSNet) is omni-scale feature learning using a residual block composed of multiple convolutional streams while a novel unified aggregation gate is introduced to dynamically fuse multi-scale features with input-dependent channel-wise weights.

We also use some pre-trained model and weights with our custom classification layer for training the characters in our PoseTReID datasets. The same to MLFN module, we use the same 30 images of full body crop of all characters, and then train with the same SVM configuration.

\subsubsection{OSNet-AIN}

OSNet-AIN \cite{osnet_ain} is pretty much like OSNet which can distinguish similar-looking people, yet generalizable for deployment across datasets without any adaptation. To improve this generalisable feature learning, the authors introduced instance normalisation (IN) layers into OSNet by optimally formulating an efficient differentiable architecture search algorithm. They achieved state-of-the-art performance yet having much smaller model than other existing reID models.

We also use the best pre-trained model and weight with our custom classification layer for training the characters in our PoseTReID datasets. The same to OSNet module, we use the same 30 images of full body crop of all characters, and then train with the same SVM configuration.

\begin{figure}[ht]
  \centering
  \includegraphics[width=\linewidth]{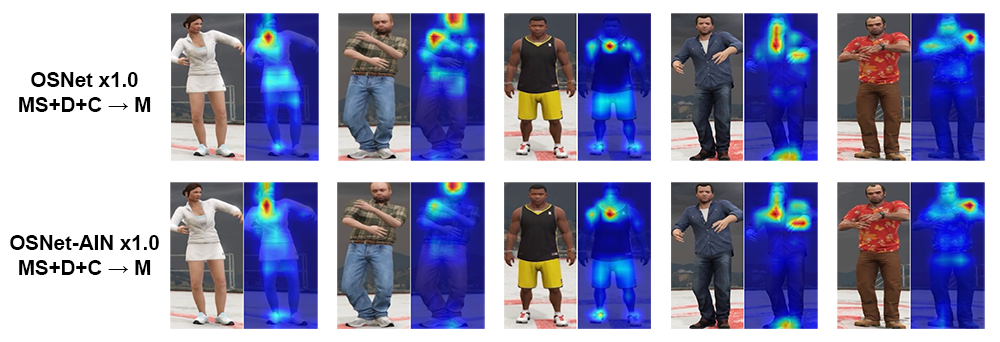}
  \caption{Feature map comparison of the 5 characters in PoseTReID datasets: OSNet x1.0 vs OSNet-AIN x1.0. Both are based on pre-trained models and Weight MS+D+C\textrightarrow M. The extracted features or areas of interest of both results are very similar. Interestingly, the region of interests of each character look unique and distinguishable, which is a good indication of re-identification.}
  \label{fig:osnet}
\end{figure}

\subsection{Datasets}

In our last paper \cite{ptreid9271712}, we explained why we used our own synthetic datasets not other popular datasets like \cite{MOT19_CVPR}\cite{MOT20}. Surprisingly soon later, we started to see booming growth in synthetic datasets such as MOTSynth-MOT-CVPR22 \cite{MOT_synth}.

To evaluate the results, we use our PoseTReID datasets simulated by a 3D video game, GTA V \cite{gta_v} which is a great tool for creating controlled environment datasets for people tracking, face and body recognition and re-identification, vehicle analysis, and especially the scenarios where real world situations are costly and hard to establish. Resolutions, duration, and more details can be found in our last paper \cite{ptreid9271712}. 

To understand the results in section \ref{sec:res} better, you may need to understand the complexities of PoseTReID datasets a bit here. The datasets have two complexities for people tracking which are illustrated in figures \ref{fig:normal} and \ref{fig:hard}. ``Normal'' complexity videos feature 3 characters who never leave but always stay in camera field of view. Two of the characters make a single occlusion by passing in front of one another. This normal complexity has only two camera positions as shown in figure \ref{fig:normal} (Left image as ``Low'' position, and right image as ``High'' position). 

\begin{figure}[ht]
  \centering
  \includegraphics[width=\linewidth]{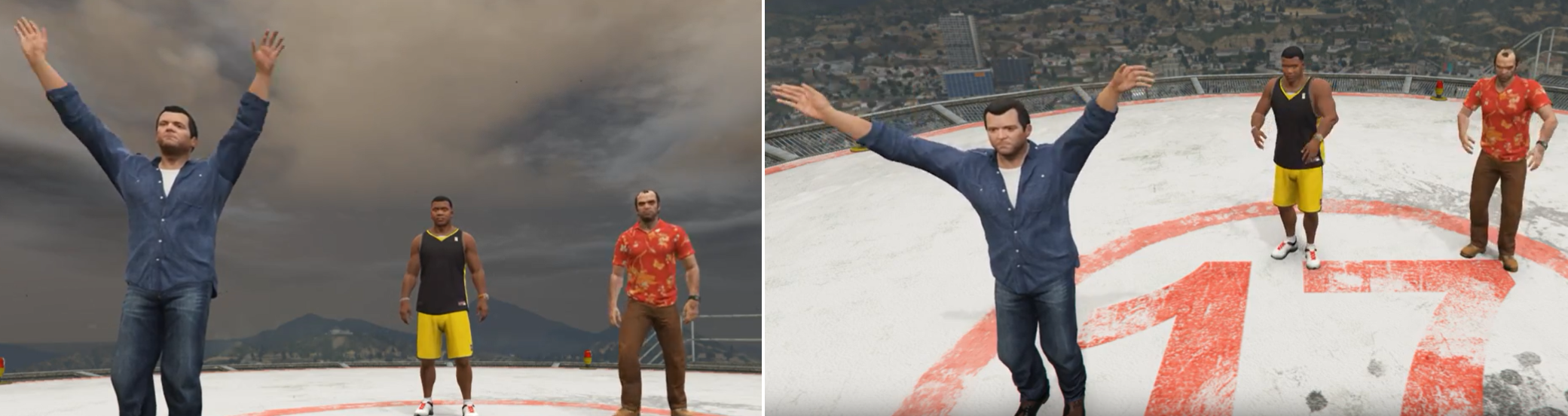}
  \caption{Two frames extracted from each of normal complexity videos, which illustrate 2 different camera position configurations: Low (Left) and High (Right). Two of the characters on the right come closer to the camera and make an occlusion by crossing each other in the later frame sequences.}
  \label{fig:normal}
\end{figure}

``Hard'' complexity videos feature 5 characters. Two of the characters leave and reappear again in the camera field of view, and they also make several occlusions on their ways in and out. This hard complexity is made by four camera position configurations as shown in \ref{fig:hard}: (1) ``Low'' position as the top-left frame, (2) ``Frontal'' position as the top-right frame which has the highest occlusion probability, (3) ``High'' position as the bottom-left, and (4) ``Surveillance'' position as the bottom-right image which is hard for face re-identification to recognize faces.  

\begin{figure}[ht]
  \centering
  \includegraphics[width=\linewidth]{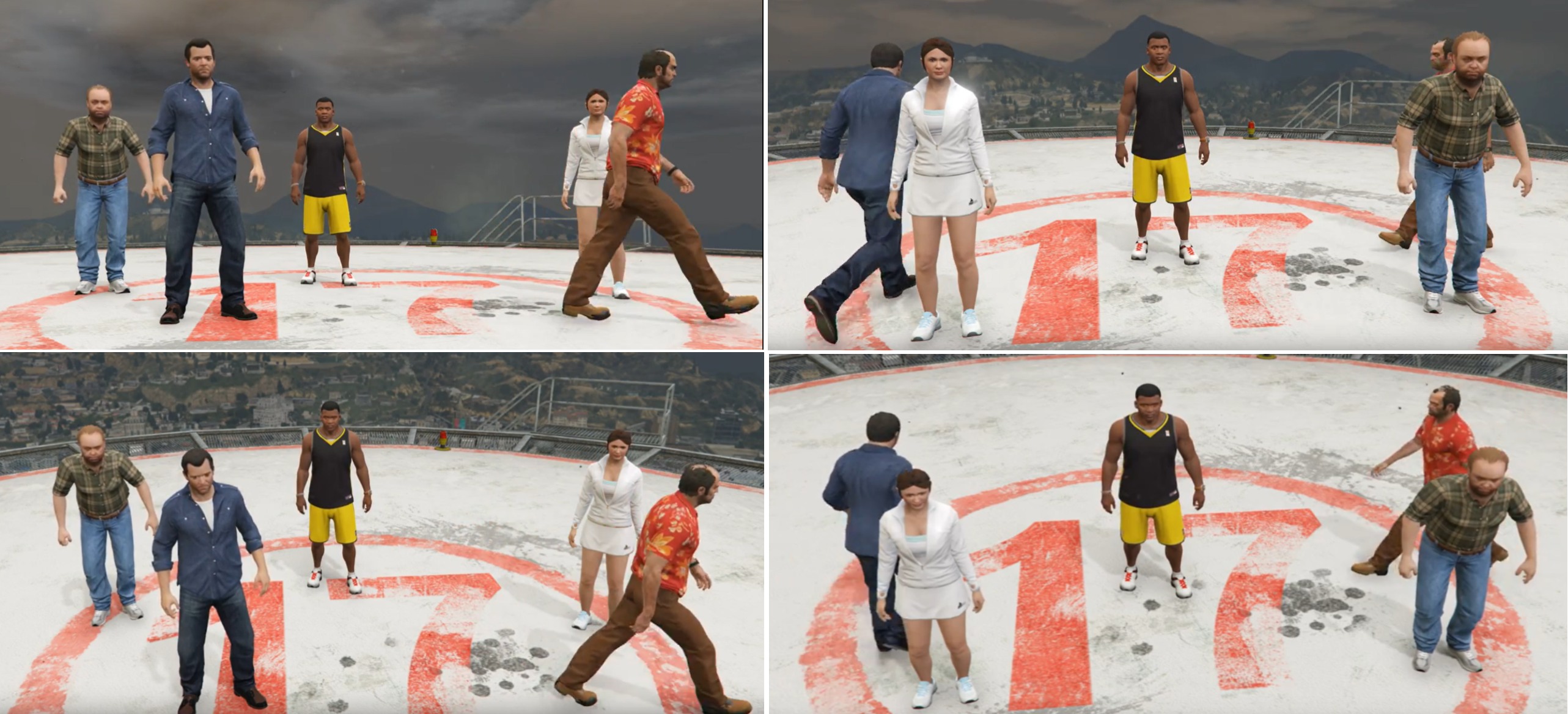}
  \caption{Four frames extracted from each of hard complexity videos, which illustrate 4 different camera position configurations: Low (Top-left), High (Bottom-left), Frontal (Top-right), and Surveillance (Bottom-right). Every configuration has 5 characters, and two of them go out of the camera field of view, make occlusions with others, and come back in the later frame sequences.}
  \label{fig:hard}
\end{figure}

\section{Quantitative Results and Discussion}
\label{sec:res}

Be noticed that our pyppbox has a layer which manages and handles various types of inputs for various types of trackers which are used in this experiment. For example, SORT and DeepSORT trackers, which require input as a set of bounding boxes from detector, can also be used together with OpenPose which is a pose or keypoints detector. To deal with this, pyppbox simply generates a bounding box from keypoints. Refer to pyppbox documentation for more details. 

Before discussing any result from the table, be noticed that, in DT (Detector) column in table \ref{tab:comp_normal} and \ref{tab:comp_hard}, ``FN'' refers to false negative meaning that the detector fails to detect a person, while ``FP'' refers to false positive meaning that the detector detects a none person as a person.

\begin{table}[!htbp]
\caption{Comparison Results: Normal-High}
\begin{center}
\begin{threeparttable}
\setlength{\tabcolsep}{1.85pt}
\begin{tabular}{|c|c|c|c|c|c|c|}
\hline
DT & Tracker & Result & ReIDer & ReID Count & Incorrect ID & Result \tabularnewline
\hline
\hline
\multirow{15}{*}{\rotatebox[origin=c]{90}{OpenPose (FN=0, FP=2)}} & 
\multirow{5}{*}{Centroid} & \multirow{5}{*}{{\bf 100\%}} 
& +FaceNet & 6 & 2 & 99.85\%\tabularnewline
& & & +MLFN & 3 & 0 & {\bf 100\%}\tabularnewline
& & & +OSNet-AIN\tnote{a}\phantom{ } & 3 & 0 & {\bf 100\%}\tabularnewline
& & & +OSNet\tnote{b} & 3 & 0 & {\bf 100\%}\tabularnewline
& & & +OSNet\tnote{c} & 3 & 0 & {\bf 100\%}\tabularnewline
\cline{2-7} \rule{0pt}{2ex}
& \multirow{5}{*}{SORT} & \multirow{5}{*}{85.27\%} 
& +FaceNet & 15 & 3 & {\bf 99.78\%}\tabularnewline
& & & +MLFN & 12 & 3 & {\bf 99.78\%}\tabularnewline
& & & +OSNet-AIN\tnote{a}\phantom{ } & 12 & 3 & {\bf 99.78\%}\tabularnewline
& & & +OSNet\tnote{b} & 12 & 3 & {\bf 99.78\%}\tabularnewline
& & & +OSNet\tnote{c} & 12 & 3 & {\bf 99.78\%}\tabularnewline
\cline{2-7} \rule{0pt}{2ex}
& \multirow{5}{*}{DeepSORT} & \multirow{5}{*}{84.54\%} 
& +FaceNet & 21 & 5 & 99.63\%\tabularnewline
& & & +MLFN & 16 & 2 & {\bf 99.85\%}\tabularnewline
& & & +OSNet-AIN\tnote{a}\phantom{ } & 16 & 2 & {\bf 99.85\%}\tabularnewline
& & & +OSNet\tnote{b} & 16 & 2 & {\bf 99.85\%}\tabularnewline
& & & +OSNet\tnote{c} & 16 & 2 & {\bf 99.85\%}\tabularnewline
\hline
\hline
\multirow{15}{*}{\rotatebox[origin=c]{90}{YOLO v4 (FN=0, FP=0)}} & 
\multirow{5}{*}{Centroid} & \multirow{5}{*}{{\bf 99.85\%}} 
& +FaceNet & 19 & 13 & 99.04\%\tabularnewline
& & & +MLFN & 3 & 2 & {\bf 99.85\%}\tabularnewline
& & & +OSNet-AIN\tnote{a}\phantom{ } & 3 & 2 & {\bf 99.85\%}\tabularnewline
& & & +OSNet\tnote{b} & 3 & 2 & {\bf 99.85\%}\tabularnewline
& & & +OSNet\tnote{c} & 3 & 2 & {\bf 99.85\%}\tabularnewline
\cline{2-7} \rule{0pt}{2ex}
& \multirow{5}{*}{SORT} & \multirow{5}{*}{{\bf 99.85\%}} 
& +FaceNet & 19 & 13 & 99.04\%\tabularnewline
& & & +MLFN & 3 & 2 & {\bf 99.85\%}\tabularnewline
& & & +OSNet-AIN\tnote{a}\phantom{ } & 3 & 2 & {\bf 99.85\%}\tabularnewline
& & & +OSNet\tnote{b} & 3 & 2 & {\bf 99.85\%}\tabularnewline
& & & +OSNet\tnote{c} & 3 & 2 & {\bf 99.85\%}\tabularnewline
\cline{2-7} \rule{0pt}{2ex}
& \multirow{5}{*}{DeepSORT} & \multirow{5}{*}{99.19\%} 
& +FaceNet & 24 & 13 & 99.04\%\tabularnewline
& & & +MLFN & 12 & 2 & {\bf 99.85\%}\tabularnewline
& & & +OSNet-AIN\tnote{a}\phantom{ } & 12 & 2 & {\bf 99.85\%}\tabularnewline
& & & +OSNet\tnote{b} & 12 & 2 & {\bf 99.85\%}\tabularnewline
& & & +OSNet\tnote{c} & 12 & 2 & {\bf 99.85\%}\tabularnewline
\hline
\end{tabular}
\begin{tablenotes}
    \footnotesize
    \item[a] OSNet-AIN: Multi-source domain generalization MS+D+C\textrightarrow M
    \item[b] OSNet: Multi-source domain generalization MS+D+C\textrightarrow M
    \item[c] OSNet: Same-domain generalization M
\end{tablenotes}
\end{threeparttable}
\end{center}
\label{tab:comp_normal}
\end{table}

Here are the details of each parameter of every module used in this experiment:
\begin{compactitem}
\item Detectors:
 \begin{compactitem}
  \item OpenPose (Model: BODY\textunderscore25, Model res: -1x256)
  \item YOLO v4 (Model res: 416x416, NMS: 0.45, Conf: 0.5)
 \end{compactitem} 
\item Trackers:
 \begin{compactitem}
  \item Centroid (Max dist: 50)
  \item SORT (Max age: 1, min hits: 3, IoU: 0.3)
  \item DeepSORT (NN: 100, NMS: 0.5, Max dist: 0.1)
 \end{compactitem} 
\item Re-Identifiers:
 \begin{compactitem}
  \item FaceNet (Min conf: 0.75)
  \item MLFN (Min conf: 0.35, Base only)
  \item OSNet-AIN x1.0 (Min conf: 0.35, MS+D+C\textrightarrow M)
  \item OSNet x1.0 (Min conf: 0.35, MS+D+C\textrightarrow M)
  \item OSNet x1.0 (Min conf: 0.35, M)
 \end{compactitem} 
\end{compactitem} 

The result in percentage (\%) is the score of having correct IDs by comparing to ground truth. Be noticed that the ground truth of our datasets was manually checked and verified to make sure that every character or person has only one unique ID across all frames of each video, unlike other datasets which use the tracking results of SORT \cite{sort} as the baseline score or the ground truth. Thus, our scoring measures how well the tracking system can keep track of the exact same person across all frames in the same video.

\begin{table}
\caption{Comparison Results: Hard-Surveillance}
\begin{center}
\begin{threeparttable}
\setlength{\tabcolsep}{1.85pt}
\begin{tabular}{|c|c|c|c|c|c|c|}
\hline
DT & Tracker & Result & ReIDer & ReID Count & Incorrect ID & Result \tabularnewline
\hline
\hline
\multirow{15}{*}{\rotatebox[origin=c]{90}{OpenPose (FN=2, FP=27)}} & 
\multirow{5}{*}{Centroid} & \multirow{5}{*}{71.77\%} 
& +FaceNet & 47 & 198 & 94.59\%\tabularnewline
& & & +MLFN & 45 & 100 & {\bf 97.26\%}\tabularnewline
& & & +OSNet-AIN\tnote{a}\phantom{ } & 45 & 101 & 97.24\%\tabularnewline
& & & +OSNet\tnote{b} & 46 & 103 & 97.18\%\tabularnewline
& & & +OSNet\tnote{c} & 36 & 101 & 97.24\%\tabularnewline
\cline{2-7} \rule{0pt}{2ex}
& \multirow{5}{*}{SORT} & \multirow{5}{*}{{\bf 72.24\%}} 
& +FaceNet & 221 & 332 & 90.92\%\tabularnewline
& & & +MLFN & 90 & 110 & {\bf 96.99\%}\tabularnewline
& & & +OSNet-AIN\tnote{a}\phantom{ } & 92 & 113 & 96.91\%\tabularnewline
& & & +OSNet\tnote{b} & 91 & 111 & 96.96\%\tabularnewline
& & & +OSNet\tnote{c} & 93 & 116 & 96.83\%\tabularnewline
\cline{2-7} \rule{0pt}{2ex}
& \multirow{5}{*}{DeepSORT} & \multirow{5}{*}{41.17\%} 
& +FaceNet & 238 & 344 & 90.60\%\tabularnewline
& & & +MLFN & 105 & 112 & 96.93\%\tabularnewline
& & & +OSNet-AIN\tnote{a}\phantom{ } & 114 & 111 & 96.96\%\tabularnewline
& & & +OSNet\tnote{b} & 113 & 108 & {\bf 97.04\%}\tabularnewline
& & & +OSNet\tnote{c} & 104 & 109 & 97.02\%\tabularnewline
\hline
\hline
\multirow{15}{*}{\rotatebox[origin=c]{90}{YOLO v4 (FN=28, FP=10)}} & 
\multirow{5}{*}{Centroid} & \multirow{5}{*}{41.85\%} 
& +FaceNet & 150 & 166 & 95.46\%\tabularnewline
& & & +MLFN & 36 & 64 & {\bf 98.25\%}\tabularnewline
& & & +OSNet-AIN\tnote{a}\phantom{ } & 43 & 76 & 97.92\%\tabularnewline
& & & +OSNet\tnote{b} & 47 & 83 & 97.73\%\tabularnewline
& & & +OSNet\tnote{c} & 44 & 78 & 97.86\%\tabularnewline
\cline{2-7} \rule{0pt}{2ex}
& \multirow{5}{*}{SORT} & \multirow{5}{*}{{\bf 41.88\%}} 
& +FaceNet & 215 & 189 & 94.83\%\tabularnewline
& & & +MLFN & 76 & 63 & 98.27\%\tabularnewline
& & & +OSNet-AIN\tnote{a}\phantom{ } & 79 & 56 & {\bf 98.46\%}\tabularnewline
& & & +OSNet\tnote{b} & 82 & 61 & 98.33\%\tabularnewline
& & & +OSNet\tnote{c} & 79 & 59 & 98.38\%\tabularnewline
\cline{2-7} \rule{0pt}{2ex}
& \multirow{5}{*}{DeepSORT} & \multirow{5}{*}{40.68\%} 
& +FaceNet & 316 & 294 & 91.96\%\tabularnewline
& & & +MLFN & 85 & 311 & 91.50\%\tabularnewline
& & & +OSNet-AIN\tnote{a}\phantom{ } & 87 & 57 & {\bf 98.44\%}\tabularnewline
& & & +OSNet\tnote{b} & 92 & 63 & 98.27\%\tabularnewline
& & & +OSNet\tnote{c} & 89 & 61 & 98.33\%\tabularnewline
\hline
\end{tabular}
\begin{tablenotes}
    \footnotesize
    \item[a] OSNet-AIN: Multi-source domain generalization MS+D+C\textrightarrow M
    \item[b] OSNet: Multi-source domain generalization MS+D+C\textrightarrow M
    \item[c] OSNet: Same-domain generalization M
\end{tablenotes}
\end{threeparttable}
\end{center}
\label{tab:comp_hard}
\end{table}

Table \ref{tab:comp_normal} shows the comparison results on ``Normal'' complexity and ``High'' camera position video which has 455 frames and 1,365 detection count (Sum of IDs in all frames) according to the ground truth. OpenPose has 0 FN but 2 FPs while YOLO v4 has 0 FN and 0 FP. For detection alone, YOLO v4 outperforms OpenPose here.

For tracker-only (Without ReIDer) results in table \ref{tab:comp_normal}, the combination of (OpenPose + Centroid) has the highest score 100\% despite OpenPose having 2 FP meaning that the false positive detection given by OpenPose here does not interfere with the actual persons that exist in the ground truth. It is obvious that ``Centroid'' with neck point can achieve 100\% here in this scenario due to the fact that the necks do not overlap each other during the occlusion. The combinations of (OpenPose + SORT) and of (OpenPose + DeepSORT) have similar results around 85\% because one of the persons or characters is assigned by a new ID after occlusion. DeepSORT is designed to deal with occlusion yet fails to do so here, which is somehow similar to yet slightly worse than SORT due to the fact that DeepSORT needs first 3 frames for initializing. ``ID switch'' does not occur in any combination of YOLO, and similar to the combination of OpenPose, DeepSORT is slightly worse than others because it needs extra frame for its initialization.

For overall tracking results with ReIDer in table \ref{tab:comp_normal}, all combinations achieve over 99\%, which are more or less the same. By applying the tweak for FaceNet as mentioned in section \ref{sssec:facenet}, the combination of (OpenPose + Centroid + FaceNet) achieves up to 99.85\% vs. 96.85\% from the same combination in our last paper, see table \ref{tab:comp_old_new}. However, by comparing to combinations with other ReIDers, each combination with FaceNet has the highest ``ReID Count'' among its own group of the same tracker, meaning that FaceNet is more struggling to re-identify the characters than other ReIDers.

\begin{table}
\caption{Comparison: Previous vs. Current works}
\begin{center}
\begin{threeparttable}
\setlength{\tabcolsep}{3.5pt}
\begin{tabular}{|c|c|c|}
\hline
Complexity & Set & Result \tabularnewline
\hline
\hline
\multirow{6}{*}{Normal-High}
& Prev. \cite{ptreid9271712} (STAF \cite{raaj2019efficient_staf}) & 85.05\%\tabularnewline
& Prev. \cite{ptreid9271712} (OpenPose + Centroid) & {\bf 100\%}\tabularnewline
& Prev. \cite{ptreid9271712} (OpenPose + Centroid + FaceNet) & 96.85\%\tabularnewline
& Curr. (OpenPose + Centroid + FaceNet) & 99.85\%\tabularnewline
& Curr. (OpenOose + DeepSORT + MFLN) & 99.85\%\tabularnewline
& Curr. (YOLO v4 + SORT + OSNet-AIN \tnote{a} ) & 99.85\%\tabularnewline
\hline
\hline
\multirow{6}{*}{Hard-Surveillance}
& Previous \cite{ptreid9271712} (STAF \cite{raaj2019efficient_staf}) & 86.22\%\tabularnewline
& Prev. \cite{ptreid9271712} (OpenPose + Centroid) & 85.96\%\tabularnewline
& Prev. \cite{ptreid9271712} (OpenPose + Centroid + FaceNet) & 40.71\%\tabularnewline
& Curr. (OpenPose + Centroid + FaceNet) & 94.59\%\tabularnewline
& Curr. (OpenOose + DeepSORT + MFLN) & 91.50\%\tabularnewline
& Curr. (YOLO v4 + SORT + OSNet-AIN \tnote{a} ) & {\bf 98.46\%}\tabularnewline
\hline
\end{tabular}
\begin{tablenotes}
    \footnotesize
    \item[a] OSNet-AIN: Multi-source domain generalization MS+D+C\textrightarrow M
\end{tablenotes}
\end{threeparttable}
\end{center}
\label{tab:comp_old_new}
\end{table}

Hard-Surveillance video used in table \ref{tab:comp_hard} has much longer duration (876 frames) and many more occlusions. There are totally 3,660 detection count (Sum of all IDs in all frames). For detector comparison here, we can say that OpenPose performs more accurate than YOLO v4 as OpenPose has only 2 vs. 28 FNs. For the results of tracker-only (Without ReIDer) here, we can say that SORT outperforms others while DeepSORT still does not perform as it is supposed to.

On Hard-Surveillance video, the combination of (OpenPose + Centroid + FaceNet) achieves up to 94.59\% vs. 40.71\% from the same combination in our last paper, see table \ref{tab:comp_old_new}, meaning the improved classification layer now is much better and it gives fewer incorrect IDs to the tracker in this scenario where the camera position is very high and not ideal for face recognition. In contrast, the tracker-only result here (OpenPose + Centroid) can only achieve 71.77\% (Table \ref{tab:comp_hard}) vs. 85.96\% in our last paper (See table \ref{tab:comp_old_new}), and this happens because in our last paper, we implemented everything in C++ (Except FaceNet) where OpenPose could run much faster even with higher model resolution (Higher model resolution gives better detection and stability). 

When it comes to Hard-Surveillance, the gaps between overall tracking performance of using FaceNet as re-identifier and using other deep re-identifiers are noticeably bigger according to table \ref{tab:comp_hard}. FaceNet is more struggling to re-identify than other re-identifiers as some characters in the video do not always turn their faces to the camera especially when they walk back in the camera field of view after they exit. This allows the whole body discriminative models to have a greater advantage over FaceNet. 

For computing power consumption and framerate (fps), we did all experiments on a laptop (Intel i7-8750H, DDR4 32 GB of RAM, and RTX 2080 Max-Q, 8 GB of VRAM) running on Microsoft Windows 11 64-bit, and the average framerate was about 12 fps for (OpenPose + Centroid/SORT), 9 fps for (OpenPose + DeepSORT), 30 fps for (YOLO v4 + Centroid/SORT), and 20 fps for (YOLO v4 + DeepSORT), regardless the re-identifier module as it is fast and only causes framerate to drop about a few frames when it is used. 

\section{Conclusion}

Our detailed comprehensive study of PoseTReID framework in this paper really shows the potential of the overall tracking performance with various types of detectors, trackers, and re-identifiers where good re-identifier does make PoseTReID framework very robust regardless the type of tracking method. On top of that, the adjustment on our custom classification layer has proved to help boost the accuracy performance of re-identifier as well as the overall tracking performance of PoseTReID framework.

We believe that our Python toolbox pyppbox is a right step forward to help research community and attract more young researchers in the related fields.

\section*{Acknowledgment}

The authors would like to gratefully thank to the supports of Belgium Wallonie Win2Wal program, under APTITUDE project.

\balance
\bibliographystyle{IEEEtran}
\bibliography{refs}

\end{document}